\documentclass[runningheads]{llncs}

\usepackage{eccv}
\usepackage[utf8]{inputenc}
\usepackage{eccvabbrv}
\usepackage{textcomp}
\usepackage{booktabs}
\usepackage{graphicx}
\usepackage{amsmath}
\usepackage[accsupp]{axessibility}  

\usepackage{hyperref}
\usepackage{orcidlink}

\usepackage[utf8]{inputenc}
\usepackage[T1]{fontenc}
\usepackage{amsmath,amssymb,amsfonts}
\usepackage{hyperref}
\usepackage{url}
\usepackage{booktabs}
\usepackage{nicefrac}
\usepackage{microtype}
\usepackage{graphicx}
\usepackage{tabularx}
\usepackage{enumitem}
\usepackage{xcolor}
\usepackage{array}
\usepackage{float}
\graphicspath{{./images/}}

\newcommand{\AH}{A/H}

\title{CALM-AH: An ABAW11-Calibrated Multimodal Ensemble with
Reliability-Gated Multi-Expert Consensus for Video-Level
Ambivalence and Hesitancy Recognition}

\title{CALM-AH: An ABAW11-Calibrated Multimodal Ensemble with
Reliability-Gated Multi-Expert Consensus for Video-Level
Ambivalence and Hesitancy Recognition}
\titlerunning{CALM-AH}

\author{
Wenzhuo Sun\inst{1}\textsuperscript{*} \and
Mingjian Liang\inst{1}\textsuperscript{*} \and
Richard Attfield\inst{1} \and
Zongyuan Ge\inst{1} \and
Xuelian Cheng\inst{2}\textsuperscript{\textdagger} \and
Pamela Carreno-Medrano\inst{1}
}
\authorrunning{W. Sun et al.}

\institute{
Monash University, Melbourne, Australia\\
\email{wenzhuo.sun@monash.edu, richard.attfield@monash.edu, 
       zgee0005@monash.edu, pamela.carreno@monash.edu}
\and
Monash Suzhou Research Institute, Suzhou, China\\
\email{Xuelian.Cheng@monash.edu}
}

\begin{document}
\maketitle

\begin{abstract}
Ambivalence and hesitancy (A/H) are subtle behavioural states that
may be expressed through language, voice, facial activity, and other
non-verbal cues. The ABAW11 A/H Video Recognition Challenge asks
systems to assign a binary A/H label to each naturalistic interview
video. Performance is measured using Macro-F1 so that recognition of
both A/H and No-A/H samples receives equal importance.

We present CALM-AH, a multimodal ensemble that combines textual,
acoustic, visual, and derived behavioural-statistical features. We
construct 15 non-empty combinations of these feature branches. For
each combination, we select the best of three classifier families using
validation binary cross-entropy and optimise its decision threshold for
validation Macro-F1. The resulting binary decisions are combined using
fixed hard-voting weights transferred from BROTHER.

We further introduce \textbf{Reliability-Gated Multi-Expert Consensus}
(RG-MEC), an anchor-preserving decision-level ensemble that combines an
initial prediction with three complementary correction experts:
CALM-AH, AffectGPT, and a GPT-based semantic verifier. The initial system
provides the default prediction. Its label is overridden only when all
three correction experts unanimously support the same alternative class;
otherwise, the anchor prediction is retained. This unanimity-gated design
limits the influence of isolated expert errors while permitting
bidirectional correction when task-specific, multimodal-affective, and
semantic-pragmatic evidence are fully consistent.

On the participant-disjoint ABAW11 dataset, CALM-AH achieves
a Macro-F1 of $0.7525$, and the complete
RG-MEC system achieves $0.7771$.
\end{abstract}    
\section{Introduction}

The 11th Affective \& Behavior Analysis in-the-Wild (ABAW11)
Workshop and Competition includes a video-level Ambivalence/Hesitancy
(A/H) Recognition Challenge~\cite{abaw11announcement}. Given a
naturalistic interview video, participating systems must predict whether
ambivalence or hesitancy is present. The task is evaluated on the BAH
dataset using participant-disjoint data partitions, making
generalisation to previously unseen individuals a central
requirement~\cite{gonzalez2026bah}.

Ambivalence refers to the coexistence of conflicting evaluations toward
the same decision or behaviour, whereas hesitancy may be expressed
through delayed responses, qualification, self-correction, or difficulty
committing to an action~\cite{macdonald2015vaccine,miller2013motivational}.
These states are relevant to motivational interviewing and other
human-centred decision-support settings, where identifying uncertainty
or conflicting intent may help characterise how a person responds to a
proposed behavioural change.

Automatic A/H recognition is challenging because the relevant evidence
is often subtle, temporally distributed, and expressed across several
modalities. A speaker may use uncertain language while displaying little
facial activity, or may make a confident verbal statement accompanied by
pauses, prosodic variation, or visible discomfort. Conversely, pauses
and self-corrections also occur during ordinary speech and should not
automatically be interpreted as A/H. Successful systems must therefore
combine linguistic, acoustic, and visual information without treating
every cross-modal disagreement as evidence of ambivalence.

Recent challenge systems have investigated conflict-aware multimodal
fusion, text-centred prediction, and heterogeneous model
committees~\cite{bekhouche2026conflict,pereira2026brother,
savchenko2025hsemotion}. These studies demonstrate the importance of
linguistic information, but reliable integration of acoustic, visual,
and behavioural evidence remains difficult, particularly under the
limited participant-level supervision available in BAH.

We present \textbf{CALM-AH}, an ABAW11-specific reconstruction of the heterogeneous multimodal ensemble introduced by BROTHER~[4]. All data-dependent components, including visual dimensionality reduction, feature normalisation, candidate classifiers, decision thresholds, and the final inference pipeline, are re-estimated using the current training and validation partitions.

CALM-AH combines F2LLM textual features, HuBERT acoustic features, SigLIP2 visual features, and a 102-dimensional behavioural-statistical representation. We train candidate classifiers for all 15 non-empty modality combinations, select one model per combination using validation binary cross-entropy, and independently calibrate its decision threshold for validation Macro F1. The resulting binary predictions are fused by fixed-weight hard voting.

For final inference, we introduce
\textbf{Reliability-Gated Multi-Expert Consensus} (RG-MEC), an
anchor-preserving decision-level ensemble comprising an initial prediction
anchor and three complementary correction experts: CALM-AH, AffectGPT,
and a GPT-based semantic-pragmatic verifier. The initial system provides
the default prediction. CALM-AH contributes task-calibrated textual,
acoustic, visual, and behavioural evidence; AffectGPT provides
multimodal-affective reasoning; and the GPT-based expert evaluates
semantic and pragmatic evidence related to uncertainty, qualification,
self-correction, and conflicting intent.

The four sources are deliberately non-exchangeable. The initial system
acts as the stability anchor, whereas the remaining three systems form a
unanimity-gated correction committee. No individual correction expert,
and no pair of correction experts, can independently change the anchor
label. An override is permitted only when CALM-AH, AffectGPT, and the
GPT-based expert unanimously predict the same class. If the three experts
disagree, RG-MEC retains the initial prediction. This design supports
bidirectional correction while limiting the propagation of isolated
expert errors.

Our contributions are threefold:
\begin{enumerate}
\item We develop CALM-AH, a calibrated heterogeneous ensemble over textual, acoustic, visual, and behavioural representations for ABAW11 video-level A/H recognition.
\item We systematically compare hard voting with continuous-margin, question-conditioned, and regularised linear fusion, showing that independently calibrated hard decisions provide more reliable generalisation.
\item We propose RG-MEC, an asymmetric inference-time consensus mechanism
that assigns non-exchangeable roles to four decision sources. The initial
system acts as the default stability anchor, while CALM-AH, AffectGPT,
and a GPT-based semantic verifier form a unanimity-gated correction
committee. The anchor is overridden only when all three correction
experts agree on the same alternative label.
\end{enumerate}
\section{Related Work}
\subsection{Ambivalence and hesitancy recognition}
Research on automated affect analysis has traditionally focused on facial expressions, action units, or continuous valence and arousal. Ambivalence and hesitancy require a broader behavioural interpretation. The same observable cue can have different meanings depending on its context: a pause may indicate uncertainty, thoughtfulness, turn-taking, or a recording artefact; a positive statement may be sincere or may coexist with visible discomfort. The BAH dataset addresses this gap by providing expert annotations of A/H together with multimodal observations and participant-level information \cite{gonzalez2026bah}.

The challenge setting encourages systems tailored to this construct rather than the direct reuse of generic emotion classifiers. Suggested research directions include temporal modelling, cross-modal alignment, specialised fusion, domain adaptation, personalisation, and parameter-efficient adaptation of multimodal foundation models \cite{abaw11announcement}. These directions are complementary, but each must be evaluated carefully because the limited sample size makes high-capacity adaptation vulnerable to overfitting.

\subsection{Multimodal affective modelling}
Multimodal machine learning commonly combines information at the feature, representation, decision, or hybrid level \cite{baltrusaitis2019multimodal}. For \AH recognition, simple agreement-based fusion may be insufficient because meaningful evidence can involve either consistency or inconsistency among modalities. At the same time, treating every disagreement as ambivalence risks false positives caused by imperfect transcripts, missing faces, background noise, or natural conversational variation.

Our framework extends decision-level multimodal fusion in two distinct
directions. At the first level, we reconstruct and recalibrate the
heterogeneous BROTHER committee using the current ABAW11 partitions.
This level bounds the influence of potentially miscalibrated MLP, RF,
and GBDT probability outputs by converting them into independently
thresholded votes before weighted fusion.

At the second level, we introduce an asymmetric multi-expert consensus
mechanism. This stage differs fundamentally from flat majority voting,
probability averaging, and conventional stacking because the participating
systems are not assigned equal decision rights. CALM-AH provides the
mandatory behavioural-evidence gate; the initial system provides a
stability anchor; and AffectGPT and ChatGPT provide complementary
verification signals. The foundation models may jointly support a
CALM-AH-positive candidate, but cannot independently override a
CALM-AH-negative decision. The contribution is therefore not simply the
addition of more predictors, but the explicit allocation of
source-dependent decision authority.

\subsection{Learning with limited participant-labelled data}
Clinical and behavioural datasets are expensive to annotate, and the number of unique participants can be more important than the number of frames. Frame-level sampling may create many highly correlated observations without adding independent subjects. Consequently, validation procedures should be participant-disjoint, preprocessing statistics should be fitted only on the training partition, and model selection should be based on held-out participants. Pretrained representations, conservative fine-tuning, regularisation, and calibrated decision thresholds are commonly useful, but their benefit must be demonstrated rather than assumed.
\section{Challenge Task and Dataset}
\label{sec:dataset}
\subsection{Video-level prediction task}
For each input video, the system produces a binary decision indicating whether ambivalence or hesitancy is present. Ambivalence and hesitancy share a single positive label in the challenge; the task does not require distinguishing between them. The prediction unit is the complete video rather than an individual frame. Frame-level annotations and timestamps may be used during training, but the official output contains one prediction per video \cite{abaw11announcement}.

The challenge permits supervised, self-supervised, domain-adaptive, personalised, zero-shot, and few-shot learning setups. Public or private pretrained models and additional datasets may be used, provided that all external resources are disclosed in the paper. This flexibility makes transparent reporting essential: comparisons are meaningful only when the supervision, external data, and model-selection procedure are clearly stated.

\subsection{BAH dataset}
The current ABAW11 labelled dataset contains 1,427 videos divided into
participant-disjoint training, validation, and public-test partitions.
The training partition contains 778 videos from 195 participants, the
validation partition contains 124 videos from 30 participants, and the
public-test partition contains 525 videos from 75 participants. An
additional private-test set contains 152 unlabeled videos. No participant
appears in more than one labelled split.

\begin{table}[t]
\centering
\caption{BAH split statistics used in the ABAW11 video-level A/H task.}
\label{tab:dataset_statistics}
\begin{tabular}{lccc}
\toprule
\textbf{Split} &
\textbf{Videos} &
\textbf{A/H present} &
\textbf{A/H absent} \\
\midrule
Training     & 778 & 385 & 393 \\
Validation   & 124 & 75  & 49  \\
Public test  & 525 & 318 & 207 \\
Private test & 152 & --  & --  \\
\bottomrule
\end{tabular}
\end{table}

The participant-wise partition is central to the evaluation protocol.
All preprocessing transformations, dimensionality-reduction components,
classifier parameters, and validation-derived thresholds are fitted
without using labels from the public-test or private-test participants.

\subsection{Evaluation protocol}
The official ranking measure is the unweighted mean of the F1 scores for the positive and negative classes:
\begin{equation}
    P = \frac{1}{2}\left(F1_{\mathrm{AH}} + F1_{\mathrm{NoAH}}\right).
    \label{eq:macro_f1}
\end{equation}
This metric gives equal importance to recognising the presence and the absence of \AH, even when the class frequencies differ. Average precision for the positive class is also reported by the organisers. In addition to the aggregate score, we report class-wise precision, recall, and F1 so that a high Macro F1 cannot conceal a strong bias toward one label.

The ABAW11 protocol provides participant-disjoint training,
validation, public-test, and private-test partitions. In this manuscript,
all quantitative comparisons are reported on the labelled public-test
set. The private-test partition is reserved for official challenge
ranking, and its performance is omitted from the anonymous submission.

The structure of RG-MEC was developed during the challenge phase,
under the organiser-permitted multi-trial submission protocol. To avoid
treating challenge-submission feedback as independent test evidence, we
freeze the resulting decision rule and evaluate it on the public-test
set. The private-test feedback is not used to update feature encoders,
downstream classifier parameters, decision thresholds, or CALM-AH
fusion weights.

The final RG-MEC-assisted submission rule is reported separately from the
controlled public-test method. Its design was informed by observations
from earlier challenge submissions and is therefore considered a
post-hoc, challenge-specific decision heuristic rather than a
validation-only model-selection result. No private-test example was used
to update feature extractors, base-model parameters, per-model thresholds,
or ensemble weights.

\begin{figure*}[t!]
  \centering
  \includegraphics[width=1\textwidth]{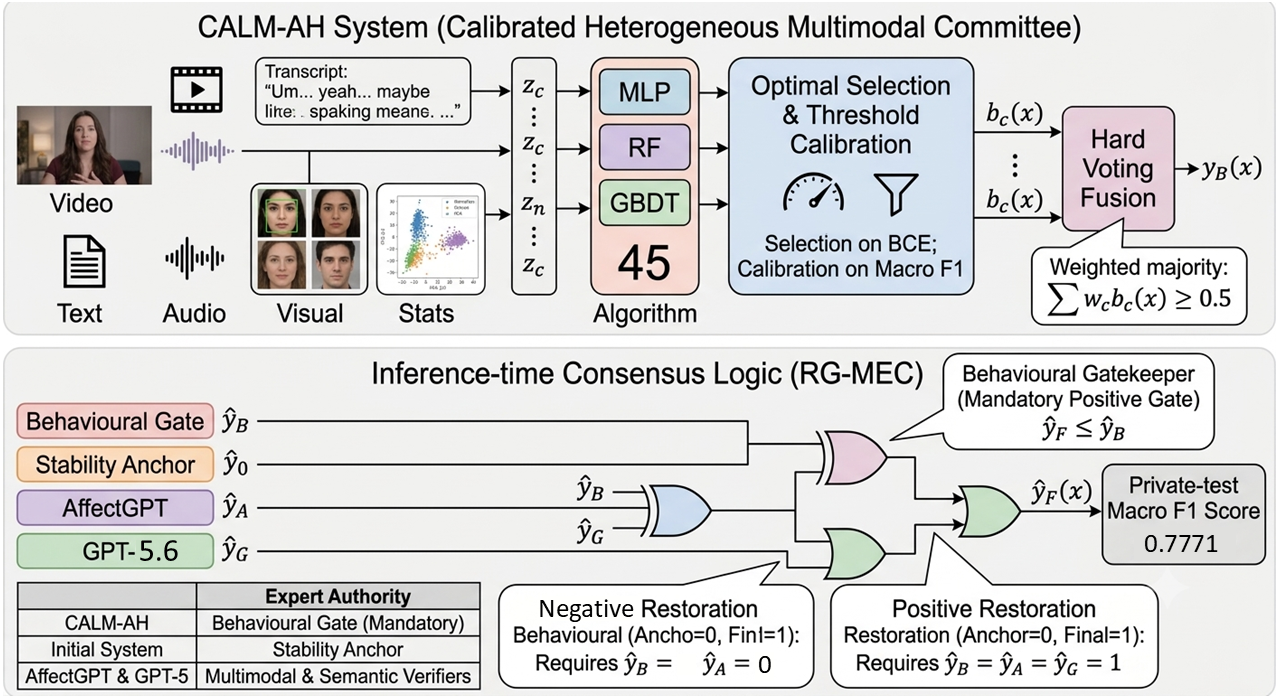}
  \caption{The pipeline of our method.}
  \label{fig:pipeline}
\end{figure*}
\section{Method}
\label{sec:method}
\subsection{Calibrated Heterogeneous Multimodal Committee}
\label{sec:committee}

Given a video $x$, we extract four complementary representations: textual semantics, speech acoustics, visual behaviour, and behavioural statistics.

For text, the complete timestamped transcript is encoded by F2LLM to preserve video-level semantic cues related to uncertainty, qualification, and self-correction. The audio stream is encoded by HuBERT to capture non-lexical information such as pauses, speaking rhythm, and prosodic variation. For the visual modality, faces are detected and aligned using RetinaFace and encoded frame-wise by SigLIP2. Invalid or temporally inconsistent frames are removed, and the remaining embeddings are aggregated with their first- and second-order temporal differences. The resulting visual descriptor is standardised and projected by PCA fitted exclusively on the training set. We further construct a behavioural-statistical representation from visual consistency, acoustic and silence statistics, transcript-level measurements, and linguistic indicators of hesitation and ambivalence.


Let $\mathcal{M}=\{T,A,V,S\}$ denote the set of four feature
branches, where $T$, $A$, and $V$ represent the primary textual,
acoustic, and visual modalities, respectively, and $S$ represents the
derived behavioural-statistical branch. We consider the set of all
non-empty branch combinations,
\begin{equation}
\mathcal{C}
=
2^{\mathcal{M}}\setminus\{\emptyset\},
\qquad
|\mathcal{C}|=15.
\end{equation}

For each modality subset, we select the classifier with the lowest validation binary cross-entropy. Its decision threshold is then independently calibrated to maximise validation Macro-F1. Model selection and threshold calibration are therefore decoupled: the former favours reliable probability estimates, while the latter determines the operating point of each classifier. This procedure yields one calibrated binary prediction for each modality subset.

All feature transformations are fitted on the training partition. Model selection and threshold calibration use only the validation partition, after which the complete committee is frozen for test inference.

\subsection{Weighted Hard Voting}
\label{sec:hard_voting}

The 15 calibrated predictions are fused using the fixed weights released with BROTHER~\cite{pereira2026brother}. Let $p_c(x)$ denote the output probability of the selected classifier for modality subset $c$, $\tau_c$ its calibrated threshold, and $w_c$ its fusion weight.


The CALM-AH prediction is

\begin{equation}
\hat{y}_{B}(x)
=
\mathbb{1}
\left[
\frac{
\sum_{c} w_c,\mathbb{1}[p_c(x)\geq\tau_c]
}{
\sum_{c} w_c
}
\geq 0.5
\right].
\label{eq:calmah}
\end{equation}

We use hard voting because probabilities produced by heterogeneous classifier families are not necessarily comparable. Independent threshold calibration maps each output to a binary decision before fusion, limiting the influence of poorly calibrated or overconfident classifiers. We retain the released weights rather than re-optimising them on the current validation partition, as the fixed weights provide additional regularisation under participant-disjoint evaluation.

\subsection{Reliability-Gated Multi-Expert Consensus}
\label{sec:rgmec}

The initial prediction and the three complementary expert systems exhibit
different error patterns. We therefore introduce
\textbf{Reliability-Gated Multi-Expert Consensus} (RG-MEC), an
anchor-preserving inference-time correction mechanism.

Let
$\hat{y}_{0}(x)\in\{0,1\}$ denote the prediction of the initial anchor
system,
$\hat{y}_{B}(x)\in\{0,1\}$ the standalone CALM-AH prediction,
$\hat{y}_{A}(x)\in\{0,1\}$ the AffectGPT prediction, and
$\hat{y}_{G}(x)\in\{0,1\}$ the prediction of the GPT-based
semantic-pragmatic verifier. Label 1 denotes A/H and label 0 denotes
No A/H.

The four decision sources are assigned non-exchangeable roles. The
initial system provides the default prediction and therefore acts as a
stability anchor. CALM-AH provides task-calibrated multimodal behavioural
evidence, AffectGPT contributes multimodal-affective reasoning, and the
GPT-based verifier contributes semantic and pragmatic reasoning related
to uncertainty, qualification, self-correction, and conflicting intent.

RG-MEC modifies the initial anchor only when all three correction experts
reach the same conclusion. We define unanimous No-A/H agreement as

\begin{equation}
u^{-}(x)
=
\neg\hat{y}_{B}(x)
\land
\neg\hat{y}_{A}(x)
\land
\neg\hat{y}_{G}(x),
\label{eq:negative_unanimity}
\end{equation}

and unanimous A/H agreement as

\begin{equation}
u^{+}(x)
=
\hat{y}_{B}(x)
\land
\hat{y}_{A}(x)
\land
\hat{y}_{G}(x).
\label{eq:positive_unanimity}
\end{equation}

A positive anchor prediction is changed to No A/H only when all three
correction experts unanimously predict No A/H. Conversely, a negative
anchor prediction is changed to A/H only when all three correction
experts unanimously predict A/H. The final prediction is

\begin{equation}
\hat{y}_{F}(x)
=
\left[
\hat{y}_{0}(x)
\land
\neg u^{-}(x)
\right]
\lor
\left[
\neg\hat{y}_{0}(x)
\land
u^{+}(x)
\right].
\label{eq:rgmec}
\end{equation}

Substituting the unanimity conditions gives the equivalent Boolean form

\begin{equation}
\begin{aligned}
\hat{y}_{F}(x)
={}&
\left[
\hat{y}_{0}(x)
\land
\left(
\hat{y}_{B}(x)
\lor
\hat{y}_{A}(x)
\lor
\hat{y}_{G}(x)
\right)
\right]
\\
&\lor
\left[
\neg\hat{y}_{0}(x)
\land
\hat{y}_{B}(x)
\land
\hat{y}_{A}(x)
\land
\hat{y}_{G}(x)
\right].
\end{aligned}
\label{eq:rgmec_boolean}
\end{equation}

For clarity, the same rule can be written in piecewise form as

\begin{equation}
\hat{y}_{F}(x)
=
\begin{cases}
0,
&
\hat{y}_{B}(x)
=
\hat{y}_{A}(x)
=
\hat{y}_{G}(x)
=
0,
\\[3pt]
1,
&
\hat{y}_{B}(x)
=
\hat{y}_{A}(x)
=
\hat{y}_{G}(x)
=
1,
\\[3pt]
\hat{y}_{0}(x),
&
\text{otherwise}.
\end{cases}
\label{eq:rgmec_piecewise}
\end{equation}

Equation~\eqref{eq:rgmec_piecewise} shows that the initial system remains
the default decision source. CALM-AH, AffectGPT, and the GPT-based expert
can override it only through unanimous agreement. Consequently, an
individual expert cannot independently modify the prediction, and
agreement between only two correction experts is also insufficient.

This formulation differs from conventional majority voting. A
two-versus-one decision among the correction experts does not determine
the final label; instead, any disagreement causes RG-MEC to retain the
initial anchor. It also differs from a backbone-veto architecture because
CALM-AH cannot independently override the anchor. RG-MEC operates only
at the decision level and leaves all feature encoders, trained
classifiers, calibrated thresholds, and hard-voting weights unchanged.
\section{Experiments}
\label{sec:experiments}

\subsection{Experimental Setup}
\label{sec:experimental_setup}

We follow the official participant-disjoint ABAW11 splits. The training partition is used to fit all feature transformations and train the 45 candidate classifiers. The validation partition is used to select one classifier for each modality combination and calibrate its decision threshold. The labelled public test set is used to evaluate the locked CALM-AH configuration and alternative fusion strategies, while the 152-video private test set is used for official challenge submission. We use video-level Macro-F1 (MF1) as the primary metric and additionally report accuracy, class-wise F1, and average precision (AP).

F2LLM, HuBERT, and SigLIP2 are used to extract textual, acoustic, and visual representations, respectively. Text and audio features are L2-normalised. The visual descriptor aggregates frame embeddings and their first- and second-order temporal differences, followed by training-set standardisation and PCA reduction to 512 dimensions. The behavioural branch contains 102 statistical descriptors derived from visual consistency, acoustic activity, silence patterns, and transcript-level hesitation cues.

For each of the 15 non-empty modality combinations, we train an MLP, a LightGBM random forest, and a LightGBM GBDT. The representative classifier is selected using validation binary cross-entropy, after which its decision threshold is calibrated for validation Macro-F1. The selected models, thresholds, preprocessing transformations, and released BROTHER fusion weights are frozen before test inference. Complete architectures, hyperparameters, feature definitions, and foundation-model inference settings are provided in the supplementary material.

No model parameter is fitted using public- or private-test labels. However, public-test performance is inspected in the post-hoc comparisons of alternative fusion and regularisation strategies. We therefore treat these comparisons as robustness analyses rather than independent model-selection experiments.

\subsection{Public Test Results}
\label{sec:public_results}



To evaluate the staged decision logic, we compare the initial anchor,
standalone CALM-AH, the conservative CALM-AH--anchor intersection, and
the complete RG-MEC configuration on the same participant-disjoint
public-test set. Standalone CALM-AH corresponds to
$\hat{y}_{B}$, while the conservative intersection corresponds to

\begin{equation}
\hat{y}_{B0}(x)
=
\hat{y}_{B}(x)
\land
\hat{y}_{0}(x).
\end{equation}

The intersection is included as an intermediate suppressive baseline and
is not identical to the complete RG-MEC rule. Full RG-MEC instead treats
the initial system as the default anchor and allows bidirectional
correction. An initial positive prediction is changed to No A/H only
when CALM-AH, AffectGPT, and the GPT-based expert unanimously predict
No A/H. An initial negative prediction is changed to A/H only when all
three correction experts unanimously predict A/H. In every other case,
the initial anchor prediction is retained.

\begin{table}[t]
\centering
\small
\caption{Public-test ablation of the asymmetric multi-expert
decision rule. All systems are evaluated on the same
participant-disjoint public-test partition.}
\label{tab:expert_ablation}
\begin{tabular}{lcccc}
\toprule
\textbf{Configuration}
& \textbf{Acc.}
& \textbf{MF1}
& \textbf{No-A/H F1}
& \textbf{A/H F1} \\
\midrule
Initial anchor $\hat{y}_0$
& 0.7790 & 0.7485 & 0.6608 & 0.8362 \\
CALM-AH $\hat{y}_B$
& 0.7752 & 0.7525 & 0.6776 & 0.8275 \\
Anchor gate $\hat{y}_B\land\hat{y}_0$
& 0.7829 & 0.7638 & 0.6968 & 0.8309 \\
Full RG-MEC $\hat{y}_F$
& 0.7981 & 0.7771 & 0.7088 & 0.8455 \\
\bottomrule
\end{tabular}
\end{table}

CALM-AH achieves a Macro-F1 of $0.7525$. Its A/H F1 is higher than its No-A/H F1, while the competitive performance on both classes indicates that the result is not obtained by simply favouring the positive class. The substantial improvement over the vision-only baseline further demonstrates the value of combining linguistic, acoustic, visual, and behavioural evidence.

\section{Conclusion}
We presented CALM-AH, an ABAW11-calibrated multimodal ensemble for
video-level ambivalence and hesitancy recognition. The system combines
textual, acoustic, visual, and behavioural-statistical representations
across 15 modality combinations, selects heterogeneous classifiers using
validation binary cross-entropy, calibrates model-specific thresholds,
and performs released-weight hard voting. The locked configuration
achieves a participant-disjoint public-test Macro F1 of $0.7525$.

We further introduced RG-MEC, an asymmetric decision-level ensemble in
which the initial prediction serves as the default stability anchor,
while CALM-AH, AffectGPT, and a GPT-based semantic verifier form a
unanimity-gated correction committee. An anchor prediction is overridden
only when all three correction experts agree on the same alternative
class; otherwise, the original anchor decision is retained. This
inference-time strategy achieves a participant-disjoint public-test
Macro F1 of $0.7771$.

The principal contribution of RG-MEC is therefore the allocation of
non-exchangeable decision roles rather than unrestricted voting among
additional predictors. Because the rule was informed by earlier official
trial behaviour, it is reported separately as a challenge-specific
refinement.
\bibliographystyle{unsrt}

\end{document}